\documentclass[10pt,twocolumn,letterpaper]{article}

\usepackage{wacv}
\usepackage{times}
\usepackage{epsfig}
\usepackage{amssymb}
\usepackage[utf8]{inputenc} 
\usepackage[T1]{fontenc}    
\usepackage{url}            
\usepackage{booktabs}       
\usepackage{amsfonts}       
\usepackage{nicefrac}       
\usepackage{microtype}      
\usepackage{xcolor}         
\usepackage{graphicx}
\usepackage{amsmath}
\usepackage{bm}
\usepackage{siunitx}
\usepackage{appendix}
\usepackage{listings}
\usepackage{algorithm}
\usepackage{placeins}

\def\wacvPaperID{1563} 

\wacvalgorithmstrack   

\wacvfinalcopy 

\ifwacvfinal
\usepackage[breaklinks=true,bookmarks=false]{hyperref}
\else
\usepackage[pagebackref=true,breaklinks=true,colorlinks,bookmarks=false]{hyperref}
\fi

\usepackage[capitalize]{cleveref}
\crefname{section}{Sec.}{Secs.}
\Crefname{section}{Section}{Sections}
\Crefname{table}{Table}{Tables}
\crefname{table}{Tab.}{Tabs.}

\begin{document}

\title{Accelerating Self-Supervised Learning via Efficient Training Strategies}

\author{%
  Mustafa Taha Koçyiğit \qquad
  Timothy M. Hospedales \qquad
  Hakan Bilen \\ \\
  University of Edinburgh, UK\\
  {\tt\small \{taha.kocyigit,t.hospedales,h.bilen\}@ed.ac.uk} \\
}

\maketitle
\thispagestyle{empty}


\maketitle

\begin{abstract}
Recently the focus of the computer vision community has shifted from expensive supervised learning towards self-supervised learning of visual representations. 
While the performance gap between supervised and self-supervised has been narrowing, the time for training self-supervised deep networks remains an order of magnitude larger than its supervised counterparts, which hinders progress, imposes carbon cost, and limits societal benefits to institutions with substantial resources. 
Motivated by these issues, this paper investigates reducing the training time of recent self-supervised methods by various model-agnostic strategies that have not been used for this problem. In particular, we study three strategies: an extendable cyclic learning rate schedule, a matching progressive augmentation magnitude and image resolutions schedule, and a hard positive mining strategy based on augmentation difficulty. We show that all three methods combined lead up to 2.7 times speed-up in the training time of several self-supervised methods while retaining comparable performance to the standard self-supervised learning setting.

\end{abstract}

\section{Introduction}


Learning representations without manual human annotations that can be successfully transferred to various downstream tasks has been a long standing goal in machine learning \cite{fukushima1982neocognitron,wiskott2002slow}. 
Self-supervised learning (SSL) aims at learning such representations discriminatively through pretext tasks such as identifying the relative position of image patches~\cite{doersch2015unsupervised} and solving jigsaw puzzles~\cite{noroozi2016unsupervised}. 
The recent success of SSL methods~\cite{he2020momentum,chen2020simple,chen2020improved} builds on contrastive learning where the representations are in a latent space invariant to various image transformations such as cropping, blurring and colour jittering.
Contrastive learned representations have been shown to obtain on par performance with their supervised counterparts when transferred to various vision tasks including image classification, object detection, semantic segmentation~\cite{chen2020big, grill2020bootstrap}, and extended to medical imaging~\cite{azizi2021big} as well as  multi-view~\cite{tian2020contrastive} and multi-modal learning~\cite{radford2021learning}.

\begin{figure}[t]
  \centering
  \includegraphics[width=1.\linewidth]{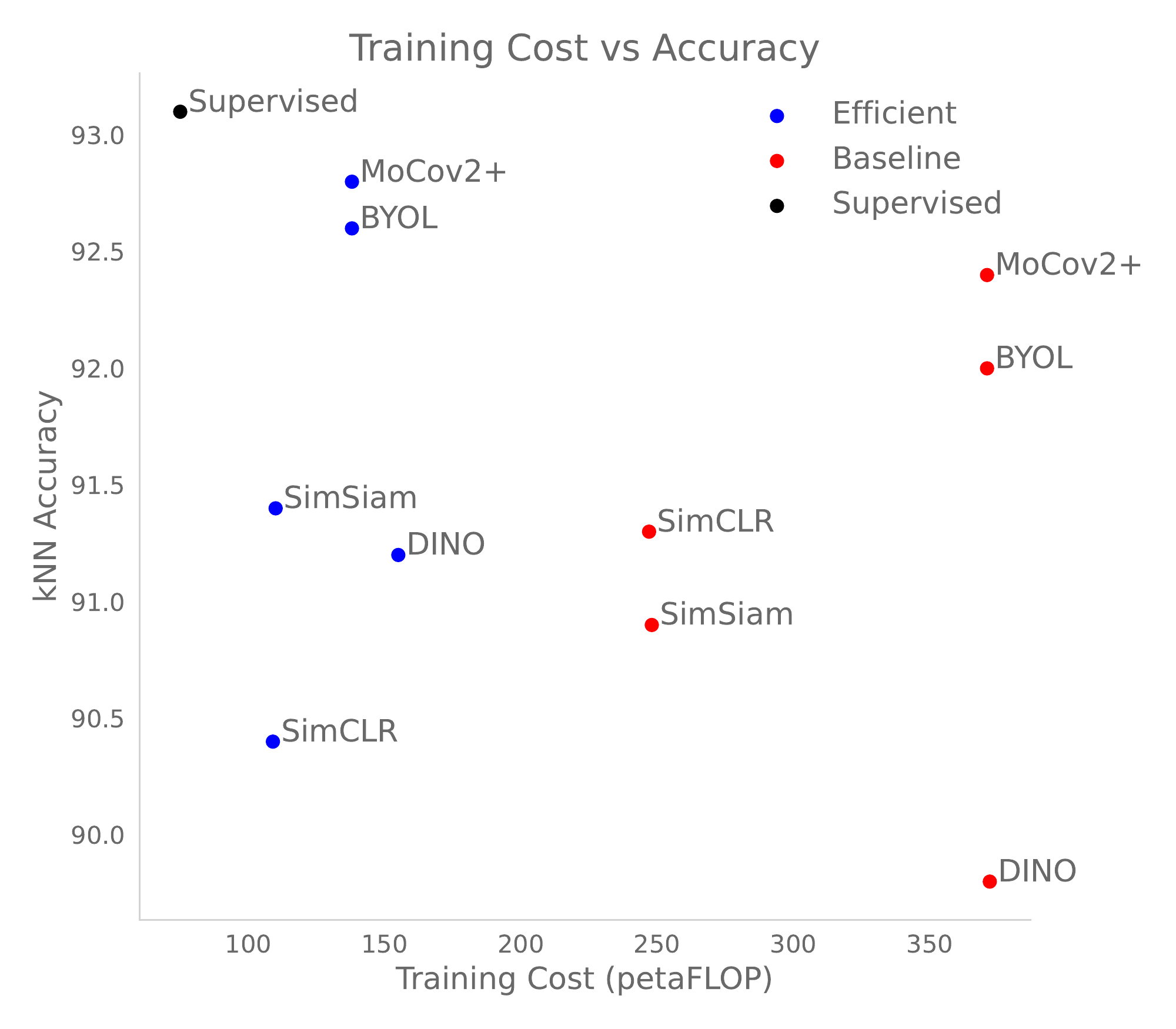}
  \caption{Classification accuracy \textit{v.s.} training cost of various SSL methods, which are trained on a subset of ImageNet by using ResNet50 (see \cref{subsec:model_ecl} for experimental details). Our method (drawn in blue) successfully significantly accelerates the SSL methods (drawn in red) without any significant drop in their performance.    }
\end{figure}

Despite the remarkable progress, an important downside of SSL methods is their high training cost which hampers the development and adoption of these promising techniques. 
Even the most efficient SSL methods require at least an order of magnitude more computation to reach the performance of supervised methods, \textit{e.g.} training BYOL \cite{grill2020bootstrap} requires 23 times more computation resulting from 8.8 times more iterations and  2.6 times more computation at each iteration than its supervised counterpart to reach similar accuracy. 
The large training cost is largely due to the challenging task of learning invariant representations over a large set of images and various augmentation transforms.
In this paper, we focus on developing algorithms that can speed up the training of SSL while maintaining their performance.


Prior work in efficient supervised training reduced the cost of training by gradually increasing the training resolution together with the augmentations magnitude using Progressive Learning \cite{tan2021efficientnetv2}. 
While augmentation is a regularization mechanism in supervised learning it is the main source of supervision in SSL and therefore plays a much more important role.
Additionally, Super Convergence \cite{smith2017super} is introduced which uses a cyclic learning rate and anti-phased momentum schedule that is used to accelerate convergence in supervised learning tasks. 
The longer duration of training makes the application of Super Convergence more difficult and using it together with Progressive Learning causes instabilities in the early stage of training.

In this work, we aim to increase the training efficiency of self-supervised training methods by optimizing the learning rate, resolution, and augmentation schedules to allow reaching the same level of performance with a smaller computational budget. 
Inspired from \cite{smith2017super} and \cite{tan2021efficientnetv2} that were used in supervised learning we propose a combined learning rate and resolution schedule for self-supervised learning. 
Additionally, we modify the augmentation strategy used in self-supervised learning for faster training. 

Our contributions are three folds.
Fixed 1-cycle Learning Rate schedules (see \cref{sec:sc}) allows Super-Convergence \cite{smith2017super} to take place for longer training times by fixing the duration of its warm-up phase while extending the decay phase and comparing it against alternative learning rate schedules used in self-supervised learning.
Super Progressive Learning (see \cref{sec:sps}) proposes a suitable Progressive Learning schedule \cite{tan2021efficientnetv2} for SSL by adding a warm-up stage with full resolution at the beginning of the resolution schedule while increasing the augmentation magnitude gradually during training so that it works well with our proposed learning rate schedule.
Hard Augment (see \cref{sec:ha}) selects a hard augmentation pair dynamically from multiple low resolution augmentations in order to boost the learning signal and regularize the training.

We show that our method accelerates the training of different self-supervised learning methods and architectures while maintaining comparable performance to standard training. We provide ablation studies for analysing the contribution of different methods and determining the optimum hyperparameters of our methods and add theoretical justifications.

\section{Related Works}

Generative models like DBN \cite{hinton2006fast}, VAE \cite{kingma2013auto} and GANs \cite{goodfellow2014generative} have been used for unsupervised representation learning however discriminative SSL have proven to be more effective. 
While early work on SSL focused on designing various pretext tasks such as solving jigsaw puzzles \cite{noroozi2016unsupervised}, colouring \cite{zhang2016colorful, larsson2017colorization, zhang2017split} and  predicting rotation \cite{gidaris2018unsupervised}, more recent techniques focused on learning augmentation invariant representation  \cite{wu2018unsupervised} \cite{tian2020makes} to replace the loss functions that require manual supervision.
Recently, Contrastive learning methods\cite{he2020momentum, chen2020simple, chen2020improved} have shown promising results for various target tasks and caused renewed interest in this area. 
However, at least an order of magnitude, more computation is required when training these methods in order to get comparable performance to supervised learning. In fact, we demonstrate our method in these recent models and show that SimSiam \cite{chen2020exploring}, BYOL \cite{grill2020bootstrap}, MoCov2 \cite{chen2020improved},  SimCLR \cite{chen2020simple} and DINO \cite{caron2021emerging} training can significantly be sped up.


By adaptation gradient noise during training large scale data parallel training has been used for increasing training speed in large deep learning methods. Using many accelerators together and increasing the total batch size training speed can be scaled up almost linearly with the number of accelerators \cite{li2020pytorch}. 
This way, supervised ImageNet training can be done in minutes \cite{smith2017don} or hours \cite{goyal2017accurate} instead of days even though the total training cost remains roughly the same.
This required adapting the hyperparameters so that gradient noise was managed properly \cite{smith2017bayesian, jastrzkebski2017three, chaudhari2018stochastic} by using the  linear relationship between the batch size and learning rate \cite{goyal2017accurate, smith2017don}. 
It was shown that training with larger learning rates had also a regularization effect \cite{jastrzkebski2017three, chaudhari2018stochastic, lewkowycz2020large} and prevented convergence to sharper minima \cite{keskar2016large}.
We use the relationship between gradient noise and learning rate schedules to optimize for training efficiency instead of speed where we synchronize the learning rate schedule with our resolution schedule to reduce the total cost of training. We use a cyclic learning rate schedule \cite{smith2017super} which combines cyclic learning rate and cyclic momentum adaptation together as our learning rate schedule due to its efficient training performance and adapt it to longer training regimes common in SSL training.

Curriculum learning \cite{bengio2009curriculum} has been used to accelerate convergence for deep learning methods and when applying it to self-supervised learning, augmentation strength and image resolution are the most relevant parameters. Image resolution directly affects accuracy \cite{hoffer2019mix, touvron2020fixing} while reducing run-time quadratically. Previous work \cite{howard2018training} on efficient supervised learning gradually increased the image resolution in the DAWN benchmark \cite{coleman2017dawnbench} to accelerate training while having slightly lower performance. 
Recently \cite{tan2021efficientnetv2} proposed gradually increasing the training image resolution while increasing augmentations strength to allow for both faster and accurate training. 
While augmentations are used for regularization in supervised learning they are the main source of supervision for recent self-supervised methods and determine task difficulty.
We build on this intuition and adapt Progressive Learning for efficient self-supervised training since in order to reduce the training cost especially in the initial phase while increasing the difficulty of the task gradually.
To the best of our knowledge having a good schedule for augmentation magnitude and its relation to resolution hasn't been examined for self-supervised learning.
    
Hard positive mining is another technique that can be used to accelerate training. Importance sampling based on sample loss has been shown to accelerate supervised learning \cite{katharopoulos2018not, jiang2019accelerating}. In object detection \cite{shrivastava2016training} selects regions with the highest loss in order to select useful positives and balance positive to negative regions effectively. \cite{gong2020maxup} applied multiple augmentations and back-propagated only the augmented sample with the maximum loss in order to improve the adversarial robustness of their method while increasing supervised performance.\cite{caron2020unsupervised} have introduced a multi-crop strategy for self-supervised learning that combines low resolution and high resolution crops in order to show increased performance while trying to keep the computational cost limited. We mine hard augmentations during training by utilizing the loss based importance sampling technique to dynamically select the most useful augmentations. Since our focus is on efficient training, we used down sampled versions of the augmented images in the selection pass to minimize the overhead.

\section{Method}

In this section we introduce the techniques used to accelerate self-supervised training. Each technique can be used to accelerate training on its own but they can be combined together in a synergistic way to enable faster training. We introduce Fixed 1-cycle Learning Rate Schedule in \cref{sec:sc}, Super Progressive Learning in \cref{sec:sps} and Hard Augment in \cref{sec:ha}.

We formulate a contrastive loss function that measures difference between pairs of images and optimizing this loss allows us to learn parameters $\theta$ for a deep neural network $f_{\theta}$ that produces similar representations for augmentations of the same image and have representation that is going to be useful for various target tasks.
Let $D$ be an unlabeled dataset consisting of $|D|$ images with resolution $r$.
We randomly sample two image transformations $\tau$ and $\bar{\tau}$ from the transformation space $T$ for each training image $\bm{x}$, apply them to $\bm{x}$ to obtain two views $\bm{v}$ and $\bm{\bar{v}}$, extract their features through the deep neural network $\bm{z}=f_{\theta}(\bm{v})$ and $\bm{\bar{z}}=f_{\theta}(\bm{\bar{v}})$ respectively.
To learn the network weights $\theta$, we minimize the loss function $L$ that represents the mismatch between two representations $\bm{z}$ and $\bm{\bar{z}}$ over the dataset and transformation space:

\begin{equation}
    \min_{\theta} \mathbb{E}_{x\sim D, (\tau,\bar{\tau}) \sim T}   L( \bm{z}, \bm{\bar{z}})
\end{equation}


We use minibatch Stochastic Gradient Descent (SGD) optimizer with momentum: 
\begin{equation}
    L^{(B)}(\theta)=  \frac{1}{|B|} \sum_{\bm{x}\sim B, (\tau,\bar{\tau})\sim T} L(\bm{z}, \bar{\bm{z}} ),
\end{equation} where the minibatch $B$ consists of $|B|$ randomly sampled images from $D$ and the loss is averaged over the samples to obtain a noisy yet unbiased estimate of the true gradient.

The update rule for $\theta$ is given as:
\begin{equation}
\begin{aligned}
    \mu_{t} & = \beta_t \mu_{t-1} - \epsilon_t \nabla_{\theta_t} L^{(B)}(\theta_t) \\
    \theta_{t+1} & = \theta_t - \mu_t
\end{aligned}
\label{eq:sgdup}
\end{equation} where $ \epsilon_t $ is the learning rate and $\beta_t$ is the momentum weighs at step $t$. 
The values for the learning rate and momentum is given by a learning rate scheduler $(\epsilon_t, \beta_t) = S(t) $.

\cref{eq:sgdup} can be interpreted as stochastic differential equation (\eg \cite{smith2017bayesian}):
\begin{equation}
\begin{aligned}
    \frac{d\theta}{dt} & = \mu, & \frac{d\mu}{dt} & = \beta_t\mu-\frac{dL}{d\theta} + \eta_t 
    \label{eqn:sgd-sde}
\end{aligned}
\end{equation} where $\eta(t) \sim \mathcal{N} (0, g\bm{F}(\theta)/|D|)$ is an additive Gaussian noise originating from the stochasticity and $\bm{F}(\theta)$ is the covariance matrix for gradients of the samples and $g$ is the ``noise scale'' which is given by $g\approx \frac{\epsilon_t |D|}{b(1-\beta_t)}$. 
For our analysis we will utilize the noise in the gradients which is proportional with the learning rate $\epsilon_t$ while being inversely proportional with batch size $b$ and $1-\beta_t$. 
This relationship plays an important role when adapting hyperparameters to different setups and for giving us a theoretically grounded understanding for learning rate schedules and their relationship with Progressive Learning.

\subsection{Fixed 1-cycle Learning Rate Schedule} \label{sec:sc}

\begin{figure*}[!htb]
    \centering
    \includegraphics[width=14cm]{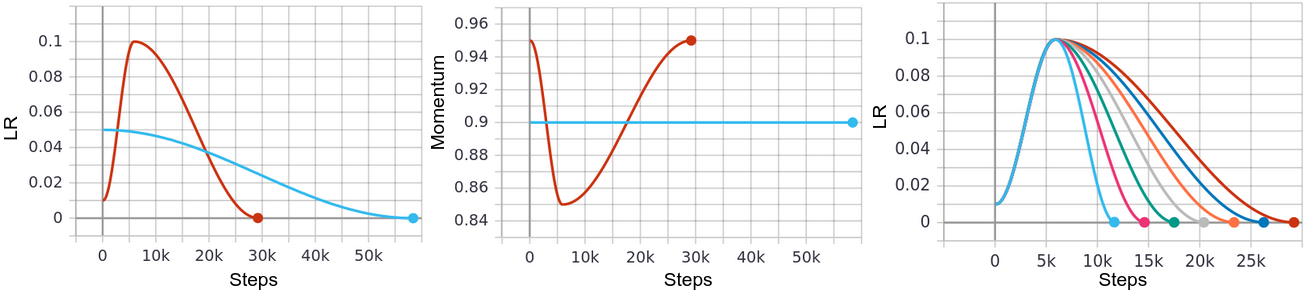}
    \caption{Learning rate (left) and momentum (middle) for standard cosine annealing learning rate schedule in blue and F1-CLR schedule in red are visualized. Learning rate for F1-CLR schedule with different length of training instances is visualized (right)}
    \label{fig:sc_lr_momentum}
\end{figure*}

Cosine annealing learning rate schedule which has been used widely in SSL \cite{he2020momentum, chen2020improved, chen2020exploring} decays the learning rate starting from a maximum learning rate value $\epsilon_{max}$ using a cosine function and can be described as, 
\begin{eqnarray}
	\epsilon_t = \frac{1}{2}\epsilon_{max}(\cos({\frac{t}{L} \pi)}+1),
    \label{eq:cosan}
\end{eqnarray} where $L$ is the total number of iterations.

Methods that perform better with larger batch sizes train with larger learning rates in order to maintain the gradient ``noise scale'' $g \propto \frac{\epsilon_t}{b}$ during training. In order to prevent instabilities in the early stages of training due to rapidly changing parameters values \cite{goyal2017accurate} a linear warm-up phase for the learning rate is added \cite{chen2020simple, grill2020bootstrap, chen2021empirical, zbontar2021barlow, caron2020unsupervised, caron2021emerging},

\begin{eqnarray}
    \label{eq:lwcosan}
    \epsilon_t= 
\begin{cases}
    \frac{t}{t_w} \epsilon_{max},   & t < t_{w} \\
    \frac{1}{2}\epsilon_{max}(\cos({\frac{t-t_w}{L-t_w} \pi)}+1),   & \text{otherwise}
\end{cases}
\end{eqnarray} where $t_w$ is the number of warm up steps.

On the other hand, Cyclic Learning Rate (CLR) schedule introduced by \cite{smith2015cyclical} and later refined as the 1-cycle learning rate (1-CLR) schedule \cite{smith2017super, smith2018disciplined} starts with a small learning rate and than increases the learning rate to a very high value and than decays it gradually to a very small value while changing the momentum at the opposite direction as the learning rate. The learning rates and momentum values at each step using a cosine window can be calculated by

\begin{eqnarray}
    \label{eq:sc}
    \epsilon_t= 
\begin{cases}
    \epsilon_{max} - \frac{1}{2}\epsilon_{max}(\cos({\frac{t}{\rho L} \pi)}+1) & t <  \rho L \\
    \frac{1}{2}\epsilon_{max}(\cos({\frac{t-\rho L}{L(1-\rho)} \pi)}+1) & \text{otherwise}
\end{cases}
\end{eqnarray}
\[
\beta_t= 
\begin{cases}
    \beta_{l}+\frac{1}{2}(\beta_{h}-\beta_{l})(\cos({\frac{t}{\rho L} \pi)}+1) & t < \rho L \\
    \beta_{h}-\frac{1}{2}(\beta_{h}-\beta_{l})(\cos({\frac{t-\rho L}{L(1-\rho)} \pi)}+1) & \text{otherwise}
\end{cases}    
\] where $\rho$ is the time percentage of time allocated for the first phase, while $\beta_l$ and $\beta_h$ are lower and higher limits for momentum respectively.
We use the learning rate (LR) range test proposed by \cite{smith2018disciplined} to set the maximum learning rate $\epsilon_{max}$  please see the details in supplementary material \ref{sec:lr_range_finder}.

A phenomenon called Super Convergence has been demonstrated in supervised classification where using larger learning rates and the 1-CLR schedule training time to reach a specified performance has been decreased dramatically \cite{howard2018training}. 
However, unlike supervised learning where longer training times do not generally result in better performance and can sometimes even cause worse performance due to over-fitting, in SSL the quality of the representation typically improves with longer training time~\cite{chen2020improved,chen2020simple}. 
A problem with the current 1-CLR is that the percentage of the first phase is being determined in proportion to the full training duration which causes an extremely long warm-up phase in longer training settings wasting compute time.
To address this problem, we propose to extend the annealing phase of 1CLR while keeping the warm up length $t_w$ the same which we call \emph{Fixed 1-cycle Learning Rate} (F1-CLR) schedule,

\begin{eqnarray}
    \label{eq:fsc}
    \epsilon_t= 
\begin{cases}
    \epsilon_{max} - \frac{1}{2}\epsilon_{max}(\cos({\frac{t}{t_w} \pi)}+1 & t < t_w \\
    \frac{1}{2}\epsilon_{max}(\cos({\frac{t-t_w}{L-t_w} \pi)}+1) & \text{otherwise}
\end{cases}
\end{eqnarray}
\[
\beta_t= 
\begin{cases}
    \beta_{l}+\frac{1}{2}(\beta_{h}-\beta_{l})(\cos({\frac{t}{t_w} \pi)}+1) & t < t_w \\
    \beta_{h}-\frac{1}{2}(\beta_{h}-\beta_{l})(\cos({\frac{t-t_w}{L-t_w} \pi)}+1).              & \text{otherwise}
\end{cases}
\]

\cref{fig:sc_lr_momentum} shows a comparison between cosine schedule and F1-CLR schedule in terms of learning rate (left) and momentum (middle) as well as illustrating the F1-CLR learning rate schedule for different lengths of training (right). 
The anti-phased movement of momentum in F1-CLR allows larger learning rates to be achieved while keeping gradient noise in check ($g \propto \epsilon_t/(1-\beta_t)$ see \cref{eqn:sgd-sde}).
Note that we do not extend the warm up phase, as the warm up is used as a stabilizer, while the learning rate and gradient noise scale are increasing ($g \propto \epsilon_t$ see \cref{eqn:sgd-sde}). 

\subsection{Super Progressive Learning} \label{sec:sps}

\begin{figure*}[t]
    \centering
    \includegraphics[width=14cm]{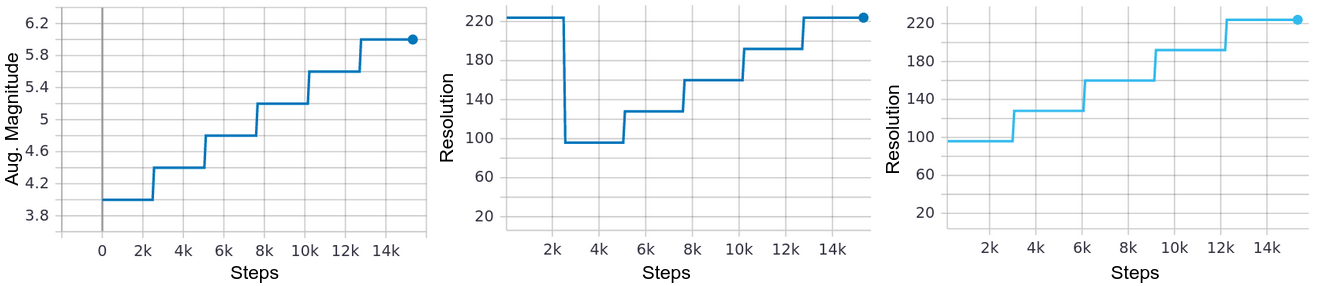}
    \caption{Augmentation magnitude and input resolution schedule where augmentation magnitude is gradually increased while the input size starts large and than gradually increases from its minimum value mimicking the inverse of the F1-CLR schedule. (left) Augmentation Magnitude, (middle) Super Progressive Learning resolution schedule, (right) Progressive Learning resolution schedule }
    \label{fig:sps_schedule}
\end{figure*}

Input resolution is very important factor that effects training time. 
Typically image resolution is a trade-off hyperparameter between performance and computational load, \textit{i.e.} higher resolution, higher performance but also more computations. 
In SSL, the standard practice \cite{he2020momentum, chen2020improved, chen2020exploring, chen2020simple, grill2020bootstrap, chen2021empirical, zbontar2021barlow, caron2020unsupervised, caron2021emerging} is to train with fixed resolution and fixed augmentation settings. 

We hypothesize that higher resolution inputs are only necessary when there is small amount of noise in the gradients, and hence, propose to employ a learning rate aware progressive learning strategy inspired by \cite{tan2021efficientnetv2}.
However, as starting training with small input resolution as in \cite{tan2021efficientnetv2} results in additional noise in the gradients which prevent it from being trivially incorporated to F1-CLR due to the instabilities in the warm-up phase.
Hence, we propose train with full resolution images during the warm-up phase before going into the linearly increasing resolution schedule while we gradually increase the augmentation magnitude as shown in \cref{fig:sps_schedule} (left, middle).
We call this strategy \emph{Super Progressive Learning}.
When the learning rate is high there is a large amount of noise in the gradients $g \propto \epsilon_t$ and this allows us to use low resolution images at that point without resulting in bad performance however since the training starts with small learning rates we need to adapt our schedule and use large resolution images.

Note that the speedup of this strategy depends on the input resolution schedule ($r_t$).
Assuming a quadratic relationship between training time and resolution, the speedup $M$ can be calculated as: 
\begin{equation}
    M = \frac{1}{L} \sum_{t} (\frac{r_{max}}{r_t})^2,
    \label{eq:sps_speedup}
\end{equation}
where $r_t$ is the input image resolution at time $t$, $r_\text{max}$ is the final resolution.
We discretize the resolution steps by 32 due to the fact that most network architectures have 5 pooling/dilation layers. The minimum value for the resolution schedule is determined empirically.

\subsection{Hard Augment} \label{sec:ha}

So far we focused on varying the effective step size and input resolution. 
Informativeness of training samples is another important factor that can speed up the training by providing more efficient gradients. 
We propose Hard Augment to boost the learning signal in self-supervised learning. We generate $m$ augmentations $\tau^m = \{ \tau_{1..m} : \tau_{1..m} \sim T\}$ for each image and evaluate the loss for all pairs of augmentations $p=\binom{\tau^m}{2}$ using a forward pass to select the augmentation pair that results in the largest amount of loss for back-propagation. 
By focusing on the samples that produce the most lost and ignoring a large fraction of the augmentations that produce a small loss and do not make a significant contribution to the training we can accelerate training dramatically. The overall objective that we optimize is 


\begin{align}\label{equ:ha} 
    L^{(B)}_{ha}(\theta)=  \frac{1}{|B|} \sum_{\bm{x} \sim D, \tau_{1..m} \sim T} \left [ \max_{ \tau_p \in p } L( \bm{z_i}, \bm{z_j})  \right].
\end{align}

\cite{gong2020maxup} have shown that using the maximum of multiple augmentations applies a regularisation on the gradient-norm  with respect to the input images $\left \| \nabla_{\bm{x}} L( \bm{z_i}, \bm{z_j}) \right\|_2$ that is in the order of $ \sigma \sqrt{\log m} $ where $\sigma$ is the strength of the augmentation under the assumption that  $\tau(x) \sim \mathcal{N} (x, \sigma^2  I)$. The regularization effect in our method can be seen as a corollary of their theorem that uses a pair of augmentations instead with a number of equivalent augmentations as the number of pairs $|p|$.


Crucially in order to decrease the overhead of selection we propose to down sample the images during the selection pass to $r_{sel}$. The image pairs that have the highest loss will than be used in training in their original resolution $r$. We do not need to have full resolution for the selection pass as we only need to have sufficient resolution to rank the augmentation pairs with respect to their loss and to find the highest loss pair which requires far less precision than is required for obtaining good quality gradient updates during training.

\begin{equation}
    M_{ha} = \frac{r^2 C}{r^2 C + mr_{sel}^2} 
    \label{eq:ha_overhead}
\end{equation}

We find the minimum resolution $r_{sel}$ that we can use for the selection pass empirically. On a typical setting in ImageNet where the original resolution is $r=224$ and the reduced selection resolution is $r_{sel}=64$ with $m=4$ augmentations and 6 possible pairs and the fraction for full training iteration cost to a forward pass cost is $C=6$ we only add $1-M_{ha}=\%5$ overhead on the training while significantly boosting the training speed by focusing on the most valuable augmentations.

\begin{algorithm}[t]
    \caption{Efficient SSL Training, PyTorch-like}
    \label{alg:code}
    \definecolor{codeblue}{rgb}{0.25,0.5,0.5}
    \definecolor{codekw}{rgb}{0.85, 0.18, 0.50}
    \lstset{
      backgroundcolor=\color{white},
      basicstyle=\fontsize{7.5pt}{7.5pt}\ttfamily\selectfont,
      columns=fullflexible,
      breaklines=true,
      captionpos=b,
      commentstyle=\fontsize{7.5pt}{7.5pt}\color{codeblue},
      keywordstyle=\fontsize{7.5pt}{7.5pt}\color{codekw},
    }
    \begin{lstlisting}[language=python]
    # f: backbone + projection layers
    # criterion: loss function for SSL method
    # aug: augmentation function
    # F1CLR: Super Convergence Schedule (Section 3.1)
    # SP: Super Progressive Schedule (Section 3.2)

    t=0 # iteration
    for e in range(epochs):
        for x in loader:  # load a minibatch x with |B| samples
           eta, beta = F1CLR(t) # Update learning rate and momentum weight 
           r, m = SP(t) # Update resolution(r) and augmentation magnitude(m) 
           vs = [aug(x, r, m) for i in range(n)]  # n random augmentation
           vi, vj = hard_augment(vs) # select hardes augmentations
           zi, zj = f(vi), f(vj) # forward-propagation
        
           L = criterion(zi,zj) # Loss calculation
           L.backward()         # back-propagation
           update(f, eta, beta) # SGD update
           t+=1
    
    def hard_augment(vs):  # negative cosine similarity
        vs_small = interpolate(vs, r_min) # down sample views to r_min
        zs = f(vs)                        # forward prop for all views
        
        indexes = arg_max_loss_pair(zs)   # determine pairs with maximum loss
        vi, vj = hard_select(vs, indexes) # select pairs by tensor indexing
        return vi, vj
    \end{lstlisting}
    \label{alg:combined}
    \end{algorithm}

We have provided the pseudo code for our method in \cref{alg:combined}. Our method can easily be adopted to different SSL methods without any major difficulty as it operates at the level of augmentations, optimizer parameters and sampling strategies.

\FloatBarrier

\begin{table*}[t]
    \centering
    \begin{tabular}{l|c|c|c|r}
& \multicolumn{2}{c|}{Baseline} &  \multicolumn{2}{c}{Efficient}  \\
        Architecture     & Acc.(\%) &  Cost (pF) & Acc.(\%) & Speedup  \\ \toprule
         ResNet18        & 51.1     &  3756     & 51.3     & 2.3x \\
         ResNet50        & 69.2     &  8494     & 69.3      & 2.3x \\
    \end{tabular}
    \caption{Our efficient strategies applied to different backbone architectures using SimSiam model on Imagenet dataset. Speed up is reported based on the reduction in training cost which is measured in terms of petaFLOPs}
    \label{tab:ecl_architecture_experiment}
\end{table*}

\begin{table*}[t]
    \centering
    \begin{tabular}{l|c|c|c|r}
& \multicolumn{2}{c|}{Baseline} & \multicolumn{2}{c}{Efficient}  \\
         Model        & Acc.(\%) &  Cost (pF) & Acc.(\%) & Speedup  \\ \toprule
         SimSiam      & 90.9     &  248       & 91.4     & 2.3x \\ 
         BYOL         & 92.0     &  371       & 92.6     & 2.7x \\
         SimCLR       & 91.3     &  247       & 90.4     & 2.3x \\
         MoCov2+      & 92.4     &  371       & 92.8     & 2.7x \\
         DINO         & 89.8     &  372       & 91.2     & 2.4x \\
    \end{tabular}
    \caption{Our efficient strategies applied to different SSL models training on ResNet50 backbone architecture on Imagenette dataset. Speed up is reported based on the reduction in training cost which is measured in terms of petaFLOPs}
    \label{tab:ecl_model_experiment}
\end{table*}

\section{Experiments}

In this section, we first study evaluate our method on different backbone architectures in \cref{subsec:backbone_ssl} and on various existing SSL methods in \cref{subsec:model_ecl}, finally we analyze the effect of each proposed component in our method in \cref{subsec:experimental_setup} and run ablation studies that analyze the hyperparameters and different trade-offs of our method in \cref{subsec:ablations_ssl}.

We evaluate the experiments on the ImageNet dataset \cite{deng2009imagenet} in terms of linear probing accuracy. This is done by freezing the backbone after pre-training and training a linear classification layer in a supervised way following \cite{chen2020exploring}. The evaluation on the Imagenette dataset\footnote{ImageNet subset that consist of 10 classes (tench, English springer, cassette player, chain saw, church, French horn, garbage truck, gas pump, golf ball, parachute) avalilable at \url{https://github.com/fastai/imagenette}} is done using online kNN accuracy of the feature maps. When calculating speed up we consider the number of steps used in training and multiply that by the number of floating point calculations made in each step and compare it against the baseline.

\paragraph{Implementation Details}
Here we provide details on the default parameters that we have used during our experiments where we use the SimSiam \cite{chen2020exploring} model with ResNet50 backbone using the SGD optimizer with .9 momentum. On ImageNet experiments we use a batch size of 512 and weight decay of 0.0001. For the baseline experiments we use the Cosine Annealing learning rate schedule with learning rate of 0.1 ($\eta_{max}$) and trained them for 200 epochs ($t_{max}$) and for our accelerated  F1-CLR schedule we use a learning rate of 0.16 ($\eta_{max}$) with 10 epoch warm-up ($t_w$) and train them for 120 epochs ($t_{max}$). For Imagenette experiments we use a batch size of 128, weight decay of 0.0005, learning rate of 0.1 ($\eta_{max}$) with Cosine Annealing scheduler and train them for 800 epochs ($t_{max}$). For our accelerated setting we use the F1-CLR schedule with learning rate of 0.2 ($\eta_{max}$) with 80 epochs for warm up ($t_w$) and train for 480 epochs ($t_{max}$). For Hard Augment we use 6 augmentations generated using SimCLR\cite{chen2020simple} transforms and scale the colour jittering linearly where the standard scale is defined as 5. For Super Progressive learning we use 96 as the minimum resolution and use 6 stages each a multiple of 32. Additional details can be seen in Supplementary Sec. \ref{sec:additional_implementation_details}.

We have observed that GPU pre-processing plays an important role in obtaining speedups in real time. Since loading images and making multiple augmentations can become a bottleneck when training with faster GPU cards like the Nvidia V100. We use the Nvidia DALI\footnote{available at \url{https://github.com/NVIDIA/DALI}} library for fast loading and augmentation of the images in the accelerators.

\subsection{Backbone Architecture Analysis}
\label{subsec:backbone_ssl}
In order to show the generality of our method to different different backbone architectures we applied our training strategies to SimSiam \cite{chen2020exploring} model on the ImageNet dataset and report linear evaluation results and speed-up in \cref{tab:ecl_architecture_experiment}. 

Our results show that our method can accelerate both ResNet-18 and ResNet-50 architectures by 2.3 times while maintaining the baseline accuracy. 

\subsection{Model Experiments} 
\label{subsec:model_ecl}

We have evaluated our acceleration strategies on different SSL models in \cref{tab:ecl_model_experiment} by pre-taining on a ResNet 50 backbone on the Imagenette dataset. We used the default hyperparameters for SimSiam\cite{chen2020exploring}, BYOL\cite{grill2020bootstrap}, MoCov2+ \cite{chen2020improved}, DINO \cite{caron2021emerging} and SimCLR \cite{chen2020simple} models while we used the same optimizer and augmentation hyper-parameters for all of them which can be seen in the supplementary Sec. \ref{sec:additional_implementation_details}.

\begin{table*}[t]
\centering
\begin{tabular}{c | c | c| l| c}
LR schedule& Sup. Prog. & Hard Aug. & kNN-Acc.(\%)& Speedup\\ \hline
CA         & -           & -          & 90.9        & 1.0x   \\
F1-CLR     & -           & -          & 90.2 (-.7)  & 1.7x   \\
CA         & -           & \checkmark & 89.8 (-1.1) & 1.8x   \\
F1-CLR     & \checkmark  & -          & 90.1 (-.8)  & 2.7x   \\
F1-CLR     & -           & \checkmark & 90.3 (-.6)  & 1.5x   \\
F1-CLR     & \checkmark  & \checkmark & 91.4 (+.5)  & 2.4x   \\
\end{tabular}
\caption{Comparing the contribution of each component in our experimental setup. Speed up is reported based on the reduction in training cost. Using SimSiam model with ResNet50 backbone architecture on Imagenette dataset. CA stands for the baseline Cosine Annealing learning rate schedule.}
\label{tab:essl_experimental_setup}
\end{table*}

We have observed that that our method can accelerate the training of different models. We note that methods with a momentum encoder typically have a higher accuracy and training cost even when trained with the same number of steps due to the additional forward passes made with the momentum encoder which helps to stabilize the training. Our method is especially effective for models with a momentum encoder where we can accelerate the training up to 2.7 times.  

\subsection{Experimental Setup}
\label{subsec:experimental_setup}

Here we compare the contribution of each component in our experimental setup in \cref{tab:essl_experimental_setup} training with ResNet50 backbone on the Imagenette dataset using default hyperparameters whenever possible. We have observed that our improvements are generally compatible with each other and allow us to train close to the same accuracy while reducing the training cost by 2.4 times. 

While Super Progressive Learning is the main driver in reducing computational cost, Hard Augment allows us to improve performance dramatically by prioritizing useful augmentations. We note that Super Progressive Learning and Hard Augment also work well energetically since Hard Augment can select from a harder set of alternatives towards the end of the training which further boost performance.

\subsection{Ablations}
\label{subsec:ablations_ssl}

In the ablation studies use the Imagenette dataset with the parameters described in the previous section. We aim to identify various parameters of our model and examine its behaviour in different circumstances.

\paragraph{Learning Rate Schedule Comparison}
\label{subsec:lr_schedule}
We compared different learning rate schedulers in isolation in order to see what percentage of improvement is attributable to the different learning rate schedulers on the ImageNet dataset in \cref{tab:sc_comparison}. We have seen that our Extended Super Convergence method still gives a reliable performance increase in this setup compared to Cosine Annealing and Cosine Annealing with linear warm up.

\begin{table}[h]
    \centering
    \begin{tabular}{l |c|c|c}
         Schedule          & Warm-Up  &  Epochs  & Acc.(\%) \\ \toprule
         Cosine Decay      & -               &  200     & 67.7    \\
         Cosine Decay      & -               &  100     & 66.1     \\
         Cosine Decay      & Linear-10       &  100     & 66.0     \\
         F1-CLR (\ref{sec:sc})  & F1-CLR (\ref{sec:sc}) &  100     & 66.4     \\
    \end{tabular}
    \caption{Comparing different learning rate schedules. Cosine decay with linear warm-up and Super convergence both have 10 epochs of warm-up in order to make the comparison more direct.}
    \label{tab:sc_comparison}
\end{table}

\paragraph{F1-CLR Warm Up} Here we make an ablation study where we train the SimSiam model for 320 epochs with different number of warm up epoch lengths in \cref{tab:sc_warmup}. We observe that 80 epochs is the optimal warm up length and we use this setting in our future experiments and ablations that utilize F1-CLR schedule. 

\begin{table}[h]
    \centering
    \begin{tabular}{l|cccccc}
         Warm Up        & 32   & 48   & 64   & 80            & 96    & 112   \\ \toprule
         Acc(\%)        & 86.8 & 86.5 & 86.9 & \textbf{87.1} & 86.8  & 86.6  \\
    \end{tabular}
    \caption{Super Convergence training where the number of warm up epochs is changed.}
    \label{tab:sc_warmup}
\end{table}


\paragraph{Minimum Resolution}


An important hyper-parameter that determines how much speed-up can be achieved with Super Progressive Learning is the minimum resolution. We made a ablation study in \cref{tab:sps_speedup} to see the maximum speedup that can be achieved with our training strategies by increasing the minimum resolution by 32 increments starting from 64. The smallest resolution that does not result in a drop in performance is 96 which we used in our other experiments while 64 was giving better speedups it wasn't maintaining the level of performance we are seeking.

\begin{table}[!h]
    \centering
    \begin{tabular}{c | c | c }
               Min. Res. & Speed up & Acc.(\%) \\ \hline
                     128 &    1.40x &  86.6    \\
                      96 &    1.60x &  86.6    \\
                      64 &    1.82x &  86.1    \\
                       0 &       3x &   N/A    \\
    \end{tabular}
    \caption{Ablation study for the Super Progressive learning speedup and accuracy trade-off with different values for minimum resolution using 224 as the maximum resolution.}
    \label{tab:sps_speedup}
\end{table}

\paragraph{F1-CLR Training Length}

\begin{table*}[t]
\begin{minipage}{.47\linewidth}
    \centering
    \includegraphics[width=0.95\linewidth]{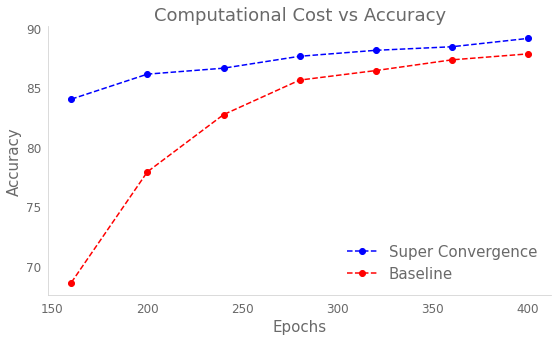}
\end{minipage}
\begin{minipage}{.47\linewidth}
\centering
        \begin{tabular}{l|c|c|c}
        \multicolumn{2}{c|}{Baseline} & \multicolumn{2}{c}{Super Convergence}  \\
         Epochs & Acc.(\%) & Speedup & Acc.(\%) \\ \hline
         160    & 68.8     & 1.61x    & 84.0     \\ 
         200    & 77.9     & 1.53x    & 86.1     \\ 
         240    & 82.7     & 1.37x    & 86.6     \\ 
         280    & 85.6     & 1.37x    & 87.6     \\ 
         320    & 86.4     & 1.42x    & 88.1     \\ 
         360    & 87.3     & 1.41x    & 88.4     \\ 
         400    & 87.8     & 1.59x    & 89.1     \\ 
    \end{tabular}
    
\end{minipage}
\caption{(left):Online validation accuracy of training instances of different lengths that are trained with F1-CLR (right): Total length of training given the fixed duration for warm-up of 80 epochs. Speedups are calculated based on linearly interpolated training length in the baseline given that the baselines has at 800 epochs has \%90.0 accuracy}
\label{fig:sc_extension}
\end{table*}

To build a better understanding of 1CLR schedule we  change the length of the training from 160 epochs to 800 epochs using the cosine annealing learning rate schedule and compare against how the proposed extended super convergence schedule performs. \cref{fig:sc_extension} shows that the difference in performance is stark when the training duration is short and gradually becomes smaller. Additionally, our F1-CLR setting leads to between 1.3x-1.6x speedup of the training to reach equivalent performance on the baseline.

\paragraph{Number of Positives}

In order to understand the effect of number of augmentations we have made an ablation study that examines the trade-off between increased overhead and the effect on accuracy  in \cref{tab:ha_positives}. We have used the SimSiam \cite{chen2020exploring} model with F1-CLR schedule and our default hyperparameters. We have observed that 6 positives gives good balance between maximizing accuracy while minimizing the overhead so we use this setting in other experiments.

\begin{table}[!h]
    \centering
    \begin{tabular}{l|cccc}
         \# of Positives & 4    & 5    & 6    & 8    \\ \toprule
          Accuracy(\%)   & 89.1 & 89.0 & 89.6 & 89.8 \\
          Overhead(\%)   & 5.4  &  6.8 &  8.2 & 10.9 \\
    \end{tabular}
    \caption{Ablation study for the numbers of positives used in Hard Augment. 6 augmentations result in good performance increase while maintaining reasonable overhead.}
    \label{tab:ha_positives}
\end{table}

\paragraph{Multi-Crop Comparison}

Multi-crop augmentation proposed by SwAV \cite{caron2020unsupervised} uses multiple augmentations with a combination of high resolution and low resolution crops where only high resolution crops are used for back-propagation and low resolution ones are used only as extra comparison targets. A slight modification to our Hard Augmentation method will produce an intermediate algorithm useful for a comparison where the additional low resolution augmentations that do not have the highest loss can be used as extra targets. We made an ablation study that compares the multi-crop augmentation strategy at the last line of \cref{tab:ha_multicrop} against our method and our method with extra targets. We train our method with Extended Super Convergence for 400 epochs and used 4 augmentations for each image.

\begin{table}[!h]
    \centering
    \begin{tabular}{c|c|c|c}
         Min Res. & Selection & Extra Target & Acc(\%) \\ \hline
         64 & Hard   &            & 89.1  \\
         64 & Hard   & \checkmark & 88.8  \\
         96 & Hard   & \checkmark & 89.7  \\
         96 & Random & \checkmark & 88.4  \\
    \end{tabular}
    \caption{Progressive learning speedup with different values for minimum to maximum resolution fraction}
    \label{tab:ha_multicrop}
\end{table}

We have seen that extra targets are useful when 96 resolution is used for the selection pass and hard augment with extra targets and hard augment outperforms multi-crop augmentation. We have also seen that the standard resolution of 64 that is enough for ranking loss pairs in Hard Augment is not enough for generating accurate targets.

Since increasing selection resolution has a direct effect on the overhead for Hard Augment and the additional improvement that can be obtained from extra targets is small we have decided not to include extra targets in our method.

\section{Conclusion}

We have demonstrated that self-supervised training can be accelerated by adapting the learning rate, augmentation and resolution schedules for self supervised training and boosting the training signal by hard positive mining on the augmentations. 
Our method have shows an training speed-up between 2.3 and 2.7 times.
which allows a much wider community to reproduce and contribute to the self-supervised learning literature, reduce the financial and carbon cost of training these models. However there is a risk that using efficient training methods the community will adopt larger and computationally more expensive benchmarks which will eliminate some of the intended benefits of our method. 
As future work we aim to apply our method to recent methods \cite{caron2021emerging,chen2021empirical} that used Vision Transformer \cite{dosovitskiy2020image} backbones.
\bibliographystyle{ieee_fullname}
\bibliography{refs}

\newpage 

\ 

\newpage
\renewcommand*{\appendixpagename}{Supplementary Material}
\appendix
\appendixpage

\section{Speed Up Calculation}

While Super Progressive Learning reduces the number of floating point calculations made by using a smaller resolution, Hard Augment adds an additional overhead for training which can be calculated by equations \ref{eq:sps_speedup} and \ref{eq:ha_overhead} respectively. Generally we find that we can reduce the number of steps by 1.7 times while reduce the number of floating point operations at each step by 1.4 times.

\section{Additional Implementation Details}
\label{sec:additional_implementation_details}


\begin{table*}[h]
    \centering
    \begin{tabular}{l|c|c|c|c}
                             & \multicolumn{2}{c|}{Imagenette} &  \multicolumn{2}{c}{ImageNet}  \\
         Hyper-parameter     & Baseline   &  Efficient & Baseline   & Efficient  \\ \hline
         Number of classes   & 10         & 10         & 1000       & 1000       \\
         Batch size          & 128        & 128        & 512        & 512        \\
         Weight decay        & \num{5e-4} & \num{5e-4} & \num{1e-4} & \num{1e-4} \\
         Learning rate       & 0.05       & 0.1        & 0.1        & 0.16       \\
         LR schedule         & CA         & ESC        & CA         & ESC        \\
         Warmup epochs       & 0          & 80         & 10         & 10         \\
         Momentum weight     & 0.9        & 0.85-0.95  & 0.9        & 0.85-0.95  \\
         Learning rate       & 0.05       & 0.05       & 0.05       & 0.16       \\
         Min aug. magnitude  & 5          & 4          & 5          & 4          \\
         Max aug. magnitude  & 5          & 6          & 5          & 6          \\
         Min view resolution & 224        & 96         & 224        & 96         \\
         Number of positives & 2          & 6          & 2          & 6          \\
         Selection resolution& N/A        & 64         & N/A        & 64         \\
    \end{tabular}
    \caption{Hyper parameters used in various setups. CA: Cosan Annealing, ESC: Extended Super Convergence}
    \label{tab:hyper_params}
\end{table*}

We have trained all our experiments in the single node setting with maximum 8 GPUs which restricted us to use 512 as the maximum batch size. Some of the SSL methods like SimCLR \cite{chen2020simple} and BYOL \cite{grill2020bootstrap} have been shown to perform better with larger batch sizes which we could not replicate and restricted ourselves to the same training setting in all our experiments.

In some of the ablations and analysis experiments where we study the effect of a certain parameter different values can be used. For Linear evaluation we use freeze the encoder and re-initialize the last layer and train with batch size 2048 and learning rate 0.8 for 90 epochs using the LARS optimizer.

Our implementation is based on PyTorch Lightning \cite{falcon2019pytorch} where we define Hard Augmentation and Super Progressive Learning Schedule as callbacks. We adapt the OneCycle Learning rate schedule implementation\footnote{\url{https://pytorch.org/docs/stable/generated/torch.optim.lr_scheduler.OneCycleLR.html}} for Extended Super Convergence. We will provide the code for our method upon publication. See the table for a full list of parameters \cref{tab:hyper_params}.

\section{Learning rate range finder}
\label{sec:lr_range_finder}
We have used the Learning rate range finder test proposed by \cite{smith2017super} to find good values for minimum and maximum learning rates. This test increase learning rate exponentially and keeps track of on the training and validation loss. The minimum value and maximum value are determined by the point at which the validation loss starts to decrease and the point at which is starts to diverge. We calculate the validation loss at each step on a batch of 4096 validation samples in order to keep the test manageable.

\begin{figure}[h]
    \centering
    \includegraphics[width=0.9\linewidth]{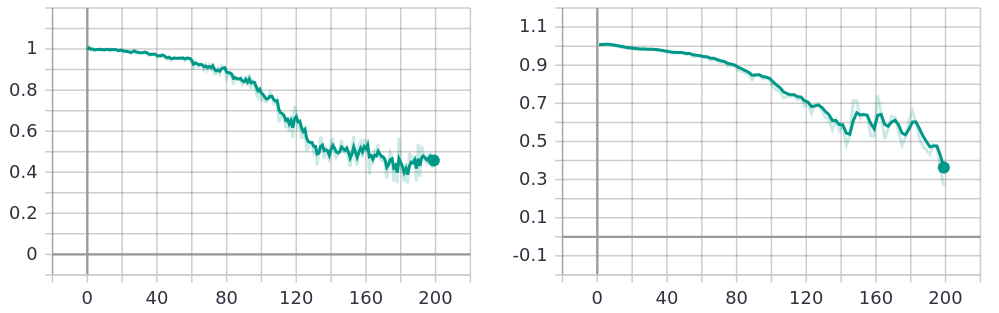}
    \caption{Learning rate finder plot for training (left) and validation (right) losses}
    \label{fig:sc_lrfinder}
\end{figure}

In the test shown in Figure \ref{fig:sc_lrfinder} we increase the learning rate from \num{1e-3} to \num{1e0} in 200 steps. Both the training and validation losses plateau close to step 135 and learning rate .1 which we use in our experiments on Imagenette dataset.

\section{Augmentation Resolution Relationship}

In order to examine the relationship between resolution and augmentation magnitude for self-supervised learning we have conducted a series of experiments. We trained a BYOL \cite{grill2020bootstrap} model on the Imagenette dataset using the ResNet18 \cite{he2015deep} architecture with RandAugment \cite{cubuk2019randaugment} augmentation with the cosine annealing learning rate schedule where we change the augmentation strength $m$ while we change the input image resolution. Each entry in the table is a separate experiment with fixed input resolution and augmentation magnitude. The results in Table \ref{tab:sps_resolution_magnitude} confirms that we can apply higher magnitude augmentations only on higher resolution images and the optimum augmentation magnitude increases with resolution. This confirms our intuition behind the relationship between resolution and augmentation magnitude in SSL and provides motivation for applying Super Progressive Learning and allows us to confirm the findings of \cite{tan2021efficientnetv2} for SSL.
\begin{table}[h]
    \centering
    \begin{tabular}{l|c|c|c|c|c}
             & m=3 &  m=5  & m=7 & m=10 & m=15 \\ \hline
         128 & 85.0 & 84.0 & 81.4 & 77.7 & 75.3 \\
         192 & 88.8 & 87.7 & 88.9 & 86.8 & 87.3 \\
         300 & 89.5 & 89.7 & 89.9 & 88.3 & 87.7 \\
    \end{tabular}
    \caption{Resolution and Augmentation Magnitude relationship shown by training a self-supervised learning method on the combination of resolution and magnitudes and measuring the online classification accuracy of a linear layer}
    \label{tab:sps_resolution_magnitude}
\end{table}

\section{Progressive Augmentation Curriculum}

However what values should be used as the minimum and maximum augmentation magnitude is an important question. Since we use the linearly scaled SimCLR \cite{chen2020simple} augmentations in our method we made a hyper-parameter search on these values in order to determine augmentation magnitudes empirically. We train our method for 320 epochs with Super Progressive learning schedule using 128 as the minimum resolution for various augmentation magnitudes and compare against the default augmentation magnitude of 5. Our experiment shows that a minimum value of 4 and maximum value of 6 perform better in our setting and outperform the fixed augmentation setting.

\begin{table}[h] 
\centering
        \begin{tabular}{c|c|c}
             $m_{min}$& $m_{max}$ & Acc(\%) \\ \hline
             5        & 5         & 88.6 \\
             2.5      & 4         & 88.4 \\
             3        & 4         & 88.7 \\
             4        & 5         & 88.7 \\
             5        & 6         & 88.7 \\
             4        & 6         & 88.9 \\
        \end{tabular}
        \caption{Maximum and minimum augmentation magnitude when trained with Super Progressive learning}
        \label{tab:sps_curriculum}
\end{table}

\section{Future Work}

This idea can be extended to other modalities where resolution is defined in different manners for example on sound with the sampling rate analogue which in a similar sense accelerates the training however than the amount of acceleration and cost benefit calculation will change. A similar case can also be made for tabular data where the most important features are processed first and later additional features are added however this is much more difficult to implement would probably require additional structure. Similarly training with a small vocabulary and than enlarging that can have a similar effect as well.

\end{document}